\documentclass[runningheads]{llncs}

\usepackage{graphicx}
\usepackage{comment}
\usepackage{amsmath,amssymb}
\usepackage{color}
\usepackage{url}
\usepackage{hyperref}

\usepackage{xcolor}
\usepackage{xr-hyper}
\usepackage{svg}
\RequirePackage{xspace}
\usepackage[inline]{enumitem}

\definecolor{mylightgreen}{RGB}{51,153,102}
\definecolor{mydarkgreen}{RGB}{17,121,74}

\definecolor{mygreenblue}{RGB}{47,111,112}
\definecolor{mylightblue}{RGB}{172,207,208} 

\definecolor{myyellow}{RGB}{215,179,18} 
\definecolor{myorange}{RGB}{215,136,18}

\definecolor{mylightred}{RGB}{192,72,72} 
\definecolor{mydarkred}{RGB}{158,28,28} 

\definecolor{mylightgray}{RGB}{238,238,238} 
\definecolor{mymedgray}{RGB}{202,204,206} 
\definecolor{mydarkgray}{RGB}{42,44,46}

\colorlet{bgcolor}{mylightgray}
\colorlet{poscolor}{mydarkgreen}
\colorlet{negcolor}{mydarkred}
\colorlet{outputscalingcolor}{blue}
\definecolor{draftcolor}{RGB}{0,73,95}
\newcommand{\isbetter}[1]{\textcolor{poscolor}{#1}}
\newcommand{\isworse}[1]{\textcolor{negcolor}{#1}}

\makeatletter
\DeclareRobustCommand\onedot{\futurelet\@let@token\@onedot}
\def\@onedot{\ifx\@let@token.\else.\null\fi\xspace}
 
\def\ie{\emph{i.e}\onedot}

\def\etal{\emph{et al}\onedot}
\makeatother

\newcommand{\methodname}{SF2SE3}
\newcommand{\norm}[1]{\left\lVert#1\right\rVert}

\newif\ifreview
\reviewtrue
\reviewfalse

\ifreview
	\usepackage{lineno}

	\linenumbers
\fi

\begin{document}


\def\SubNumber{15}

\def\GCPRTrack{Young Researcher's Forum}

\title{Sparse Dynamic Rigid Point Clouds}
\title{\methodname{}: Clustering Scene Flow into SE(3)-Motions via Proposal and Selection}

\ifreview
\else

	\author{Leonhard Sommer \and
	Philipp Schröppel \and
	Thomas Brox}
	\authorrunning{L. Sommer et al.}
	\institute{University of Freiburg, Freiburg im Breisgau, Germany \\
	\email{\{sommerl, schroepp, brox\}@cs.uni-freiburg.de} \\
	\url{https://lmb.informatik.uni-freiburg.de}
	}
\fi

\maketitle              

\begin{abstract}


We propose \methodname{}, a novel approach to estimate scene dynamics in form of a segmentation into independently moving rigid objects and their $SE(3)$-motions. \methodname{} operates on two consecutive stereo or RGB-D images. First, noisy scene flow is obtained by application of existing optical flow and depth estimation algorithms. 
\methodname{} then iteratively \begin{enumerate*}[label=(\arabic*)]\item samples pixel sets to compute $SE(3)$-motion proposals, and \item selects the best $SE(3)$-motion proposal with respect to a maximum coverage formulation\end{enumerate*}. Finally, objects are formed by assigning pixels uniquely to the selected $SE(3)$-motions based on consistency with the input scene flow and spatial proximity.

The main novelties are a more informed strategy for the sampling of motion proposals and a maximum coverage formulation for the proposal selection.
We conduct evaluations on multiple datasets regarding application of \methodname{} for scene flow estimation, object segmentation and visual odometry. \methodname{} performs on par with the state of the art for scene flow estimation and is more accurate for segmentation and odometry. 


\end{abstract}

\section{Introduction}

Knowledge about dynamically moving objects is valuable for many intelligent systems. 
This is the case for systems that take a passive role as in augmented reality or are capable of acting as in robot navigation and object manipulation. 



\begin{figure}
    \centering
    \includegraphics[scale=0.083]{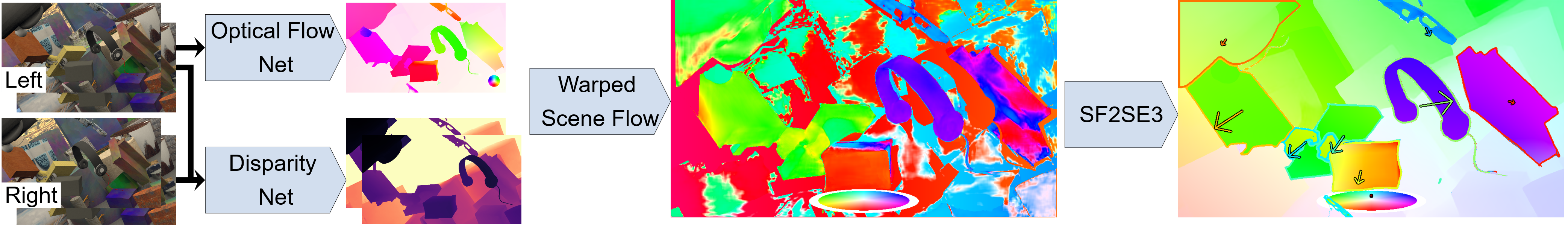}
    \caption{\textbf{Overview of \methodname{}}: Optical flow and disparity are estimated with off-the-shelf networks and combined to an initial scene flow estimate. The resulting scene flow is noisy due to inaccurate estimations and due to occlusions (center image). \methodname{} optimizes a set of objects with corresponding $SE(3)$-motions and an assignment from pixels to objects such that the initial scene flow is best covered with a minimal number of objects. Finally, an improved scene flow estimate can be derived from the object segmentation and $SE(3)$-motions, as shown in the last image. Furthermore, \methodname{} obtains the camera egomotion by determining the $SE(3)$-motion of the background.}
    \label{fig:teaser_drpc}
\end{figure}

In this work, we propose a novel approach for this task that we term Scene-Flow-To-$SE(3)$ (\methodname{}).
\methodname{} estimates scene dynamics in form of a segmentation of the scene into independently moving objects and the $SE(3)$-motion for each object. \methodname{} operates on two consecutive stereo or RGB-D images. First, off-the-shelf optical flow and disparity estimation algorithms are applied to obtain optical flow between the two timesteps and depth maps for each timestep. The predictions are combined to obtain scene flow. Note that the obtained scene flow is noisy, especially in case of occlusions. \methodname{} then iteratively \begin{enumerate*}[label=(\arabic*)]\item samples pixel sets to compute $SE(3)$-motion proposals, and \item selects the best $SE(3)$-motion proposal with respect to a maximum coverage formulation\end{enumerate*}. Finally, objects are created for the selected $SE(3)$-motions by grouping pixels based on concistency of the input scene flow with the object $SE(3)$-motion. Further, \methodname{} derives scene flow and the camera egomotion from the segmentation and $SE(3)$-motions. The described pipeline is illustrated in Fig~\ref{fig:teaser_drpc}.



Regarding related work, \methodname{} is most similar to ACOSF \cite{li2021two_acosf} in that it approaches the problem as iteratively finding $SE(3)$-motion proposals and optimizing an assignment of pixels to the proposals. However, \methodname{} introduces several improvements. 
Firstly, instead of randomly accumulating clusters that serve to estimate $SE(3)$-motion proposals, we propose a more informed strategy that exploits a rigidity constraint, \ie forms clusters from points that have fixed 3D distances. Secondly, we propose a coverage problem formulation for selecting the best motion proposal such that \begin{enumerate*}[label=(\alph*)]\item the input scene flow of all data points is best covered, and \item irrelevant or similar $SE(3)$-motions are prohibited\end{enumerate*}. The iterative process ends when no proposal fulfills the side-constraints from (b), whereas ACOSF iteratively selects a fixed number of $SE(3)$-motions.

We evaluate \methodname{} and compare to state-of-the-art approaches for multiple tasks and datasets: we evaluate scene flow estimation on the KITTI~\cite{menze2015object_kitti} and FlyingThings3D~\cite{mayer2016large_flyingthings} datasets, moving object segmentation on FlyingThings3D, and visual odometry on FlyingThings3D, KITTI and TUM RGB-D~\cite{sturm2012benchmark_tumrgbd}. We use the state-of-the-art approaches CamLiFlow~\cite{liu2021camliflow}, RAFT-3D~\cite{teed2021raft3d}, RigidMask~\cite{yang2021learning_rigidmask} and ACOSF as strong baselines for the comparison. CamLiFlow, RAFT-3D and RigidMask are currently the best approaches on the KITTI leaderboard and ACOSF is the most similar baseline to \methodname{}. 

Compared to RAFT-3D, \methodname{} obtains similar scene flow outlier rates on KITTI (\isbetter{$-0.45 \%$}) and FlythingThings3D (\isworse{$+0.47\%$}). However, the advantage is additional output information in form of the object segmentation as opposed to pixel-wise motions.

Compared to RigidMask, the scene flow outlier rate of \methodname{} is slightly worse on KITTI (\isworse{$+0.43\%$}) but significantly better on FlyingThings3D (\isbetter{$-6.76\%$}), which is due to assumptions about blob-like object shapes within RigidMask. Regarding object segmentation, \methodname{} achieves higher accuracy than RigidMask on FlyingThings3D (\isbetter{$+2.59\%$}).

Compared to ACOSF, \methodname{} decreases the scene flow outlier rate on KITTI (\isbetter{$-2.58\%$}). Further, the runtime is decrease from $5$ minutes to $2.84$ seconds. 

 On all evaluated dynamic sequences of the TUM and Bonn RGB-D dataset, \methodname{} achieves, compared to the two-frame based solutions RigidMask and VO-SF, the best performance.

To summarize, \methodname{} is useful to retrieve a compressed representation of scene dynamics in form of an accurate segmentation of moving rigid objects, their corresponding $SE(3)$-motions, and the camera egomotion.

\section{Related Work}



In the literature, different models for the scene dynamics between two frames exist: \begin{enumerate*}[label=(\arabic*)] \item non-rigid models estimate pointwise scene flow or $SE(3)$-transformations, and \item object-rigid models try to cluster the scene into rigid objects and estimate one $SE(3)$-transformation per object\end{enumerate*}. Regarding the output, all models allow to derive pointwise 3D motion. Object-rigid models additionally provide a segmentation of the scene into independently moving objects. Furthermore, if an object is detected as static background, odometry information can be derived. 

Our work falls in the object-rigid model category, but we compare to strong baselines from both categories. In the following, we give an overview of related works that estimate scene dynamics with such models.



\subsection{Non-Rigid Models}

Non-rigid models make no assumptions about rigidity and estimate the motion of each point in the scene individually as scene flow or $SE(3)$-transformations. 

The pioneering work of Vedula~\etal~\cite{vedula1999three} introduced the notion of scene flow and proposed algorithms for computing scene flow from optical flow depending on additional surface information. 

Following that, multiple works built on the variational formulation for optical flow estimation from Horn and Schunk \cite{horn1981determining} and adapted it for scene flow estimation~\cite{huguet2007variational,wedel2008efficient,basha2010multi,valgaerts2010joint,vogel20113d,jaimez2015primal}. 

With the success of deep learning on classification tasks and with the availability of large synthetic datasets like Sintel~\cite{butler2012naturalistic_sintel} and FlyingThings3D~\cite{mayer2016large_flyingthings}, deep learning models for the estimation of pointwise scene dynamics have been proposed~\cite{jiang2019sense,liu2019flownet3d,hur2020self,yang2020upgrading_oflowexp,schuster2021deep,teed2021raft3d}. In particular, in this work, we compare with RAFT-3D~\cite{teed2021raft3d}, which estimates pixel-wise $SE(3)$-motions from RGB-D images. RAFT-3D iteratively estimates scene flow residuals and a soft grouping of pixels with similar 3D motion. 
In each iteration, the residuals and the soft grouping are used to optimize pixel-wise $SE(3)$-motions such that the scene flow residuals for the respective pixel and for grouped pixels are minimized.

\subsection{Object-Rigid Models}

Object-rigid models segment the scene into a set of rigid objects and estimate a $SE(3)$-transformation for each object. The advantages compared to non-rigid models are a more compressed representation of information, and the availability of the object segmentation. The disadvantage is that scene dynamics cannot be correctly represented in case that the rigidity assumption is violated. 

Classical approaches in this category are PRSM~\cite{vogel2013piecewise_prsm,vogel20153d_prsm} and OSF~\cite{menze2015object_kitti}, which split the scene into rigid planes based on a superpixelization and assign $SE(3)$-motions to each plane. MC-Flow~\cite{jaimez2015motion_mcflow} estimates scene flow from RGB-D images by optimizing a set of clusters with corresponding $SE(3)$-motions and a soft assignment of pixels to the clusters. A follow-up work~\cite{jaimez2017fast} splits the clusters into static background and dynamic objects and estimates odometry and dynamic object motion separately. 


An early learned object-rigid approach is ISF~\cite{behl2017bounding_isf}, which builds on OSF but employs deep networks that estimate an instance segmentation and object coordinates for each instance. The later approach DRISF~\cite{ma2019deep} employs deep networks to estimate optical flow, disparity and instance segmentation and then optimizes a $SE(3)$-motion per instance such that is consistent with the other quantities.

In this work, we compare to the more recent learned approaches ACOSF~\cite{li2021two_acosf} and RigidMask~\cite{yang2021learning_rigidmask}. ACOSF takes a similar approach as OSF but employs deep networks to estimate optical flow and disparity. RigidMask employs deep networks to estimate depth and optical flow and to segment static background and dynamic rigid objects. Based on the segmentation, $SE(3)$ motions are fit for the camera egomotion and the motion of all objects. A key difference between RigidMask and our approach is that RigidMask represents objects with polar coordinates, which is problematic for objects with complex structures. In contrast, our approach takes no assumptions about object shapes.

\section{Approach}

In the following, we describe the proposed \methodname{} approach. \methodname{} takes two RGB-D images from consecutive timestamps $\tau_1$ and $\tau_2$ and the associated optical flow as input. While the optical flow is retrieved with RAFT~\cite{teed2020raft}, the depth is retrieved with a depth camera or LEAStereo~\cite{cheng2020hierarchical} in the case of a stereo camera. Using the first RGB-D image as reference image, the corresponding depth at $\tau_2$ is obtained by backward warping the second depth image according to the optical flow. Further, occlusions are estimated by applying an absolute limit on the optical flow forward-backward inconsistency~\cite{sundaram2010dense}. The depth is indicated as unreliable in case of invalid measurements and additionally for the depth at $\tau_2$ in case of temporal occlusions. \methodname{} then operates on the set $\mathcal{D}$ of all image points of the reference image:
\begin{gather}
    \mathcal{D} = \{ \: D_i = (\lefteqn{\underbrace{\phantom{x_{i}, y_{i}, z_{i}, p_i^{\tau_1}, p_i^{\tau_2}, r_{i}^{\tau_1}}}_{spatial}} x_{i}, y_{i}, z_{i}, \overbrace{ p_i^{\tau_1}, p_i^{\tau_2}, r_{i}^{\tau_1}, u_{i}, v_{i}, d_{i}, r_{i}^{\tau_2}}^{motion}) \: \},
\end{gather}
where each image point $D$ consists of its pixel coordinates $(x, y)$, its depth $z$ at $\tau_1$, its 3D points $(p^{\tau_1}, p^{\tau_2})$, its optical flow $(u, v)$, its warped disparity $d$ at $\tau_2$ and its depth reliability indications $r^{\tau_1}$ and $r^{\tau_2}$. 


The objective then is to estimate a collection of objects $\mathcal{O}$ where each object $O$ consists of a point cloud $\mathcal{P}$ and a $SE(3)$-motion $(R, t)$:
\begin{gather}
\mathcal{O} = \{ \: O_k = (\underbrace{\mathcal{P}_k}_{spatial}, \overbrace{R_k, t_k}^{motion}) \: \}.
\end{gather}

\subsection{Algorithm Outline}

\begin{figure}
    \centering
    \includegraphics[scale=0.113]{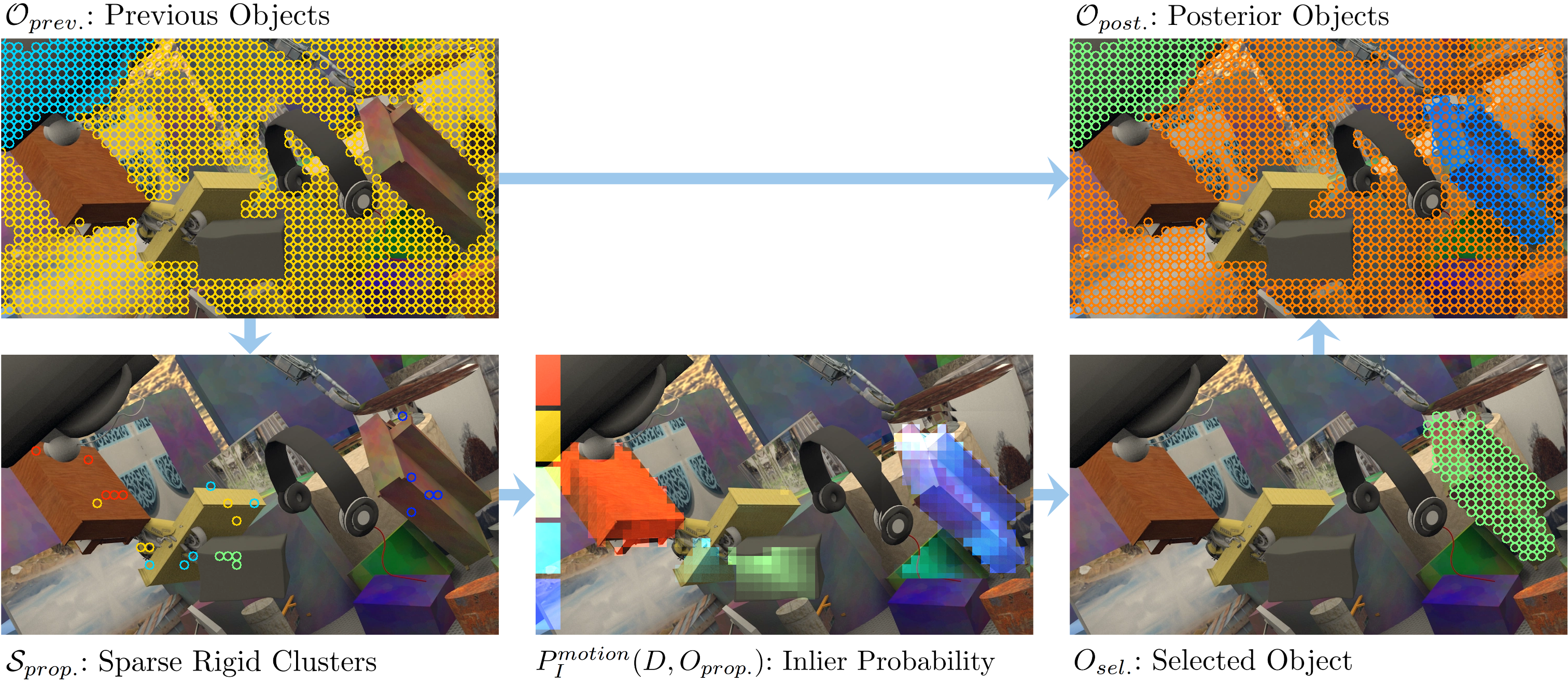}
    \caption{\textbf{Cycle of Single Object Estimation}: \methodname{} estimates objects $\mathcal{O}$ iteratively. All image points which are not covered by any previous objects $\mathcal{O}_{prev.}$, and which depth is reliable, are accumulated to obtain sparse rigid clustered $\mathcal{S}_{prop.}$. This is described in Section~\ref{sec:proposals}. Fitting an $SE(3)$-motion to each cluster results in the proposed objects $\mathcal{O}_{prop.}$ which do not contain a spatial model. Based on the inlier probabilities for the proposed and the previous objects, $P_I^{motion}(D, O_{prop.})$ and $P_I(D, O_{prev.})$, the one object is selected which maximizes the coverage objective. The coverage problem is described in Section~\ref{sec:selection}. After selecting a single object $O_{sel.}$, the image points which are covered based on the motion model $P_I^{motion}(D, O_{sel})$ are forming the point cloud which serves from then on as spatial model. Further, the selected object is split into multiple objects by splitting the point cloud into multiple spatially connected point clouds. This is not illustrated in the diagram for the sake of clarity. In case no proposed object has a sufficiently high coverage, the iterative process ends. }
    \label{fig:algorithm_outline}
\end{figure}

\methodname{} aims to estimate objects $\mathcal{O}$ which explain or rather cover the downsampled image points $\mathcal{D}$. To quantify the coverage of an image point $D$ by an object $O$, we introduce a motion inlier model $P_I^{motion}(D, O)$ and a spatial inlier model $P_I^{spatial}(D, O)$. These models are described in detail in Section~\ref{sec:inliermodels}. Based upon these models, objects $\mathcal{O}$ are retrieved iteratively, see Figure \ref{fig:algorithm_outline}.


Finally, based on the obtained rigid objects $\mathcal{O}$, \methodname{} derives odometry, segmentation and scene flow, which is described in Section~\ref{sec:deduction_odo_seg_sflow}. For this, one object is determined as background and each image point is assigned to one object based on the likelihood $f(D|O)$, which is introduced in Section~\ref{sec:inliermodels}. 

For further implementation details, including parameter settings, we publish the source code at \url{https://www.github.com/lmb-freiburg/sf2se3}.  


\subsection{Consensus Models}
\label{sec:inliermodels}

To quantify the motion consensus between the scene flow of a data point and the $SE(3)$-motion of an object, we define a motion inlier probability $P_I^{motion}(D, O)$ and likelihood $f^{motion}(D|O)$. These are defined by separately imposing Gaussian models on the deviation of the data point's optical flow in x- and y-direction and the disparity $d$ of the second time point from the respective projections  $\pi_{u}, \pi_{v}, \pi_{d}$ of the forward transformed 3D point $p^{\tau_1}$ according to the object's rotation $R$ and translation $t$. Formally, this can be written as
\begin{gather}
    \Delta u = u - \pi_{u}(R p^{\tau_1} + t) \sim \mathcal{N}(0, \sigma_{u}^2) \\
    \Delta v = v - \pi_{v}(R p^{\tau_1} + t) \sim \mathcal{N}(0, \sigma_{v}^2) \\
    \Delta d = d - \pi_d(R p^{\tau_1} + t) \sim \mathcal{N}(0, \sigma_{d}^2).
\end{gather}


Spatial proximity of the data point and the object's point cloud is measured with the likelihood $f^{spatial}(D|O)$ and the inlier probability $P_I^{spatial}(D, O)$. Therefore, we separately impose Gaussian models on the x-, y-, and z- deviation of the data point's 3D point $p^{\tau_1}$ from its nearest neighbor inside the object's point cloud $\mathcal{P}$. More precisely, we define the models
\begin{gather}
    \Delta x = x - x^{NN} \sim \mathcal{N}(0, \sigma_{geo-2D}^2) \\
    \Delta y = y - y^{NN} \sim \mathcal{N}(0, \sigma_{geo-2D}^2) \\
    \Delta z_{rel} = \frac{z - z^{NN}}{\frac{z + z^{NN}}{2}} \sim \mathcal{N}(0, \sigma_{geo-depth-rel}^2).
\end{gather}

The joint inlier probability for the spatial model yields
\begin{gather}
    P_{I}^{spat.}(D, O) = 
    \begin{cases}
    P_{I, Gauss.}(\Delta x) P_{I, Gauss.}(\Delta y) P_{I, Gauss.}(\Delta z_{rel}) &,  r^{\tau_1} = 1\\
    P_{I, Gauss.}(\Delta x) P_{I, Gauss.}(\Delta y) &,  r^{\tau_1} = 0
    \end{cases}
    \label{eq:spatial_model_inlier_prob},
\end{gather}
likewise the spatial likelihood $f^{spatial}(D|O)$ is calculated. Details for the calculation of the Gaussian inlier probability $P_{I, Gauss.}$ are provided in the Supplementary.

Regarding the motion model, the joint inlier probability yields
\begin{gather}
    P_I^{mot.}(D, O) = 
    \begin{cases}
        P_{I, Gauss.}(\Delta u) P_{I, Gauss.}(\Delta v) P_{I,Gauss.}(\Delta d) &, r^{\tau_1}=1,  r^{\tau_2} = 1 \\
        P_{I, Gauss.}(\Delta u) P_{I, Gauss.}(\Delta v) &, r^{\tau_1}=1,  r^{\tau_2} = 0 \\
        1 &, else,
    \end{cases}
\end{gather}
the same applies for the motion likelihood $f^{motion}(D|O)$.

Joining the motion and the spatial model, under the assumption of independence, results in
\begin{gather}
    f(D|O) = f^{spatial}(D|O) f^{motion}(D|O) \\
    P_I(D, O) = P_I^{spatial}(D, O) P_I^{motion}(D, O). 
\end{gather}

\subsection{Proposals via Rigidity Constraint}
\label{sec:proposals}
\begin{figure}
    \centering
    \includegraphics[scale=1.55]{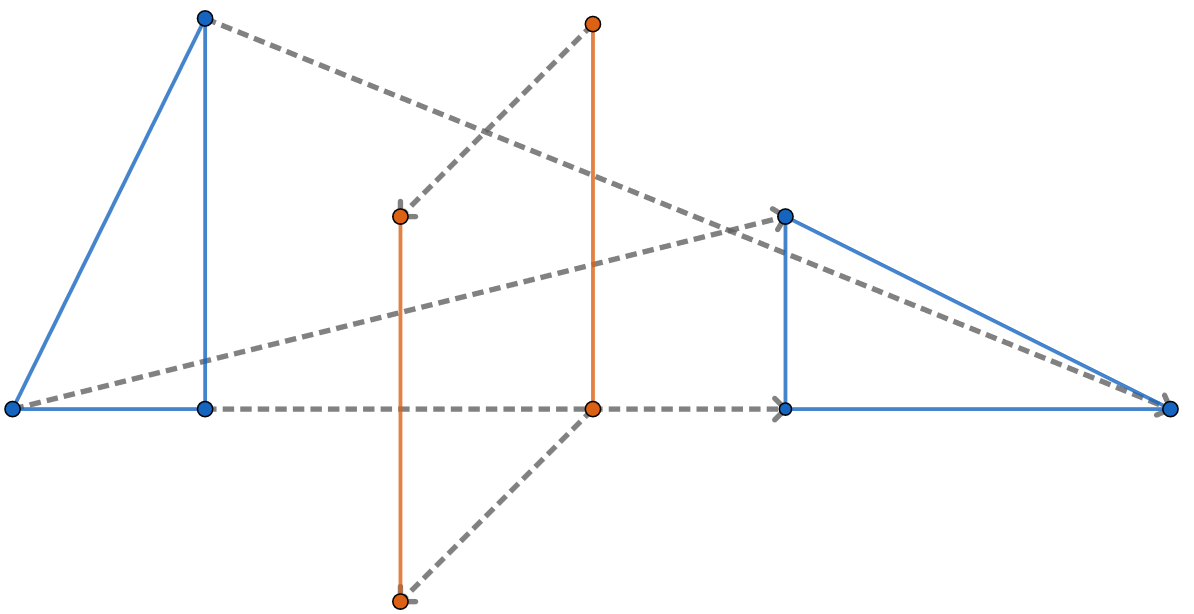}
    \caption{\textbf{Rigidity}: Points of the same color are rigid, which means that the distance between each pair remains constant despite the movement. Two accumulated points are already sufficient to calculate new $SE(3)$-motion proposals.}
    \label{fig:proposal_via_rigidity}
\end{figure}

Each proposed object is found by fitting an $SE(3)$-motion to a cluster of scene flow points which fulfill the rigidity constraint, see Figure \ref{fig:proposal_via_rigidity}. For each pair of points $(p_i, p_j)$ it must hold
\begin{gather}
     | \norm{p_{i}^{\tau_2} - p_{j}^{\tau_2}} - \norm{p_{i}^{\tau_1} - p_{j}^{\tau_1}} |  < \delta_{rigid-dev-max}.
\end{gather}
Clusters are instantiated by single scene flow points, which are sampled uniformly. Further points are iteratively added while preserving rigidity. 
Even though two points are already sufficient to estimate a $SE(3)$-motion, additional points serve robustness against noise.

\subsection{Selection via Coverage Problem}
\label{sec:selection}

Having obtained the $SE(3)$-motion proposals $\mathcal{O}_{prop.}$, we select the one which covers the most scene flow points which are not sufficiently covered by previous objects $\mathcal{O}_{prev.}$. Coverage is measured for the proposed objects with the motion model $P_I^{motion}(D, O)$. For previously selected objects, the spatial model is available in form of a point cloud. This allows us to use the joint model $P_I(D, O)$, consisting of motion and spatial model. Formally, we define the objective as
\begin{gather}
    \max_{O \in \mathcal{O}_{prop.}}\frac{1}{|\mathcal{D}|} \sum_{D \in \mathcal{D}} \max \left[ P_I^{motion}(D, O), \max_{\tilde{O} \in \mathcal{O}_{prev.}} P_I(D, \tilde{O})\right].
\end{gather}
Separating the previous coverage results in
\begin{gather}
    \max_{O \in \mathcal{O}_{prop.}} P_{contribute}(\mathcal{D}, O, \mathcal{O}_{prev.}) + \frac{1}{|\mathcal{D}|} \sum_{D_i \in \mathcal{D}} \max_{\tilde{O} \in \mathcal{O}_{prev.}} P_I(D, \tilde{O})),
\end{gather}
with the contribution probability $P_{contribute}(\mathcal{D}, O, \mathcal{O}_{prev.})$ defined as 
\begin{gather}
    P_{contribute}(\mathcal{D}, O, \mathcal{O}_{prev.}) = \frac{1}{|\mathcal{D}|} \sum_{D \in \mathcal{D}} \max \left[P_I^{motion}(D, O) - \max_{ \tilde{O} \in \mathcal{O}_{prev.}} P_I(D, \tilde{O}), 0\right].
\end{gather}
To exclude irrelevant objects, we impose for each object a minimum contribution probability.

Moreover, to exclude duplicated objects, we impose for each pair of objects a maximum overlap probability, which we define as 
\begin{gather}
    P_{overlap}(\mathcal{D}, O_1, O_2) =  \frac{ \sum_{D \in \mathcal{D}} P_I(D, O_1) P_I(D, O_2)}{ \sum_{D \in \mathcal{D}} P_I(D, O_1) + P_I(D, O_2)  - P_I(D, O_1) P_I(D, O_2)}.
\end{gather}
This overlap probability constitutes an extension of the intersection-over-union metric for soft assignments, e.g., probabilities.

Taken together, we formulate the optimization problem as
\begin{subequations}
    \label{eq:opt}
    \begin{alignat}{2}
    \max_{O \subseteq \mathcal{O}_{prop.}} \: & P_{contrib.}(\mathcal{D}, O, \mathcal{O}_{prev.})  \label{eq:opt_prob} \\
    \text{ subject to } & P_{contrib.}(\mathcal{D}, O, \mathcal{O}_{prev.}) \geq  \delta_{contrib.-min}  
    \label{eq:opt_constr1} 
    &  \\
    & P_{overlap}(\mathcal{D}, O, O_{prev.}) \leq \delta_{overlap-max}
    \label{eq:opt_constr2} 
    & \quad \forall O_{prev.} \in \mathcal{O}_{prev}.
    \end{alignat}
\end{subequations}

\begin{figure}
    \centering
    \includegraphics[scale=0.95]{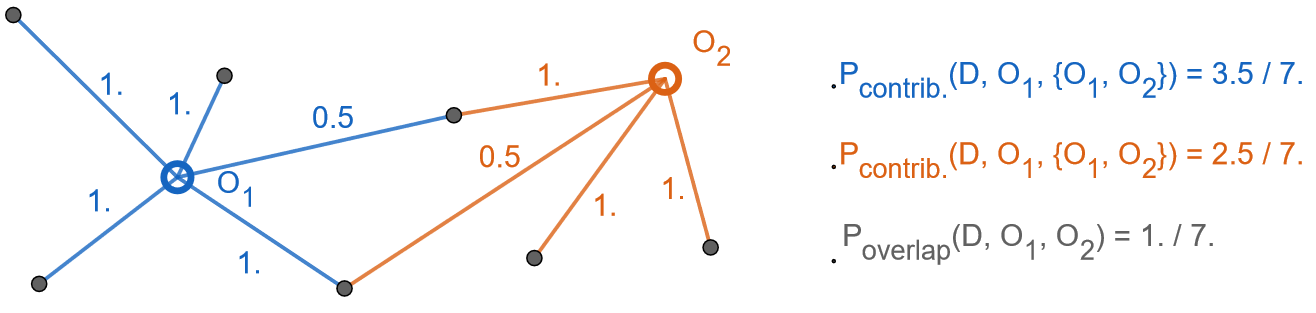}
    \caption{\textbf{Contribution and Overlap Probability for two Objects}: The edge weights equal the inlier probabilities $P_I(D, O)$, a missing edge indicates $P_I(D, O)=0$. Corresponding probabilities for contribution of each object and the overlap probability of both objects are calculated on the right side. }
    \label{fig:selection_via_coverage}
\end{figure}

An example for the calculation of contribution as well as overlap probability is provided in Figure \ref{fig:selection_via_coverage}.

Automatically the algorithm ends when the contribution probability falls below the minimum requirement.

\subsection{Deduction of Odometry, Image Segmentation, and Scene Flow}
\label{sec:deduction_odo_seg_sflow}
For estimating the odometry we determine the dynamic rigid object that equals the background. Assuming that the background equals the largest object, we choose the one that yields the largest contribution probability
\begin{gather}
    O_{background} = \arg\max_{O \in \mathcal{O}} P_{contribute}(\mathcal{D}, O, \mathcal{O}).
\end{gather}

Based on the maximum likelihood, we assign each pixel from the high-resolution image to one of the objects
\begin{gather}
    \phi_{D, O} = 
    \begin{cases}
        1 & , \arg \max_{O_k}f(D|O_k) = O \\
        0 & 
    \end{cases}.
\end{gather}

Given the object assignment the scene flow $s$ can be retrieved for each 3D point $p^{\tau_1}$ as
\begin{gather}
    s = R p^{\tau_1} + t - p^{\tau_1}.
\end{gather}

\section{Results}

We compare the performance of our method against the state of the art regarding scene flow, segmentation, odometry, and runtime.

\begin{figure}
    \centering
    \includegraphics[scale=0.0710955]{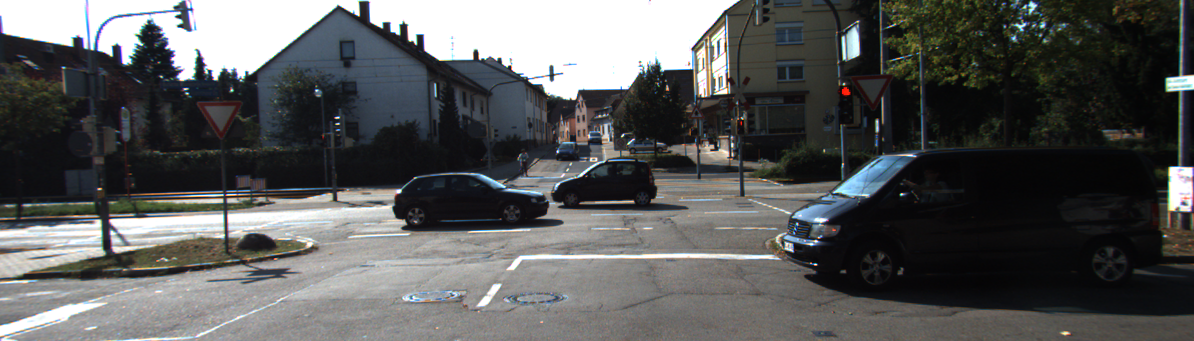}
    \includegraphics[scale=0.0710955]{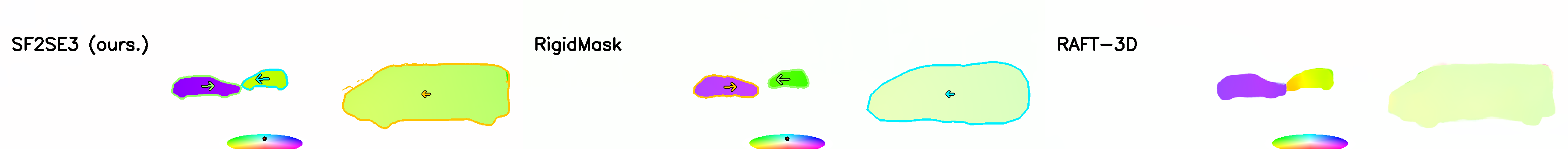}
    \includegraphics[scale=0.090122]{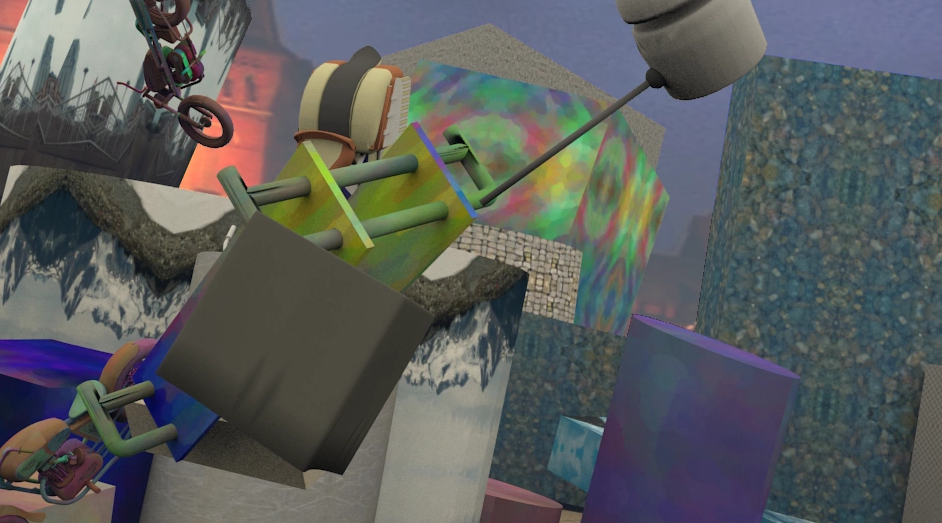}
    \includegraphics[scale=0.090122]{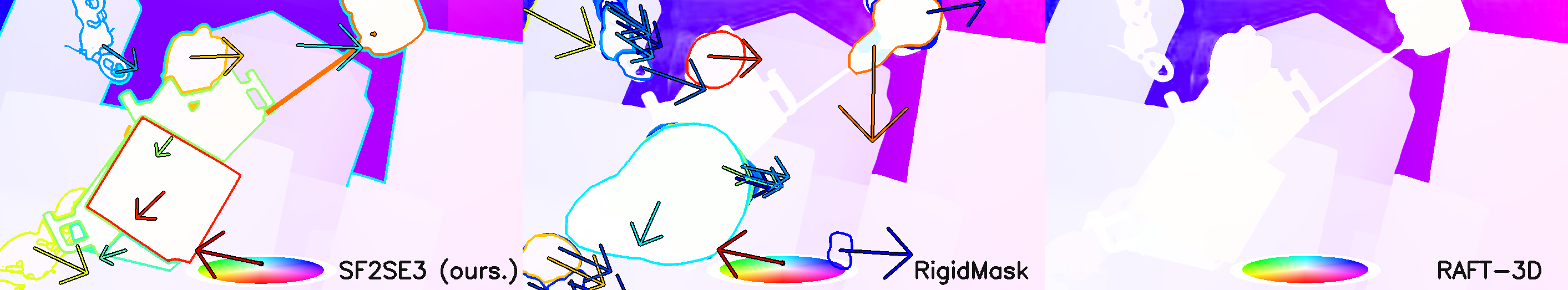}
    \includegraphics[scale=0.1401875]{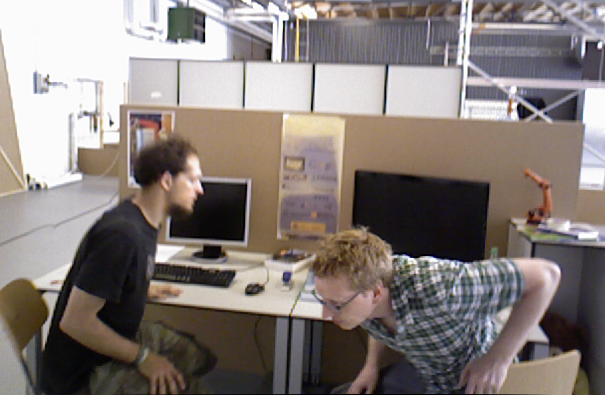}
    \includegraphics[scale=0.1401875]{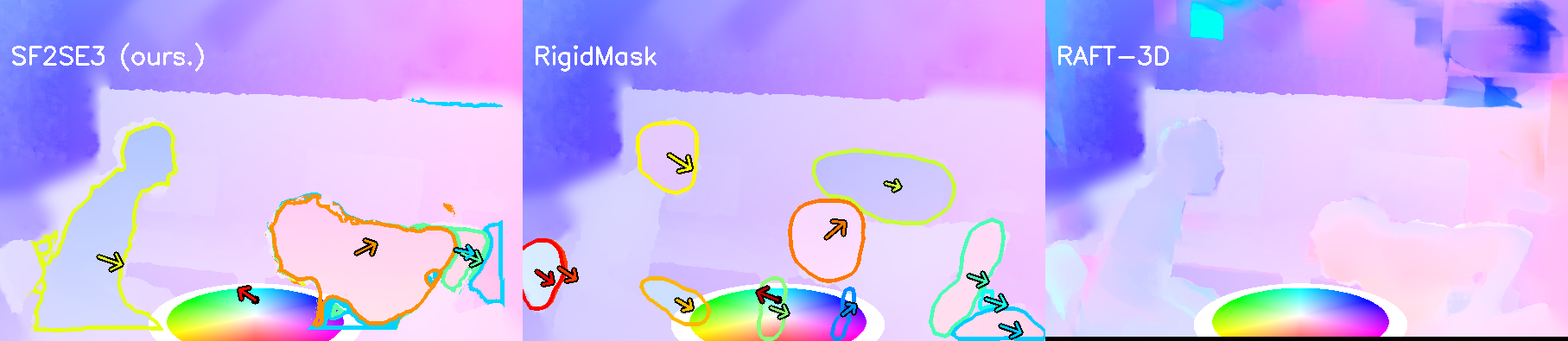}
    \caption{Qualitative results of our approach, RigidMask, and Raft-3D (left to right) on KITTI-2015, FlyingThings3D, and TUM RGB-D (top to bottom). The scene flow is color coded for the x- and z-directions, as indicated by the color wheel. The edges of the object segmentation are highlighted and the motions of object centroids are indicated with arrows. The odometry is indicated with an arrow starting from the center of the color wheel. Note that RAFT-3D estimates no segmentation and odometry.}
    \label{fig:results_samples}
\end{figure}

\paragraph{Scene Flow}

To evaluate the performance of estimating scene flow, we measure the outlier percentages of disparity, optical flow and scene flow, in the same way as the KITTI-2015 benchmark \cite{menze2015object_kitti}. The results for KITTI-2015 and FlyingThings3D are listed in Table \ref{tab:eval_sflow_outlier}.

\begin{table}[]
    \centering
        \caption{Listed are the outlier percentages for disparities at both timestamps, optical flow, and scene flow. An outlier for optical flow and disparity implies a deviation from the ground truth of $>3 \text{ pxl}$ absolutely and $>5\%$ relatively. An outlier for scene flow implies an outlier for either disparity or optical flow.}
    \label{tab:eval_sflow_outlier}
\begin{tabular}{llrrrr}
\hline
 Method            & Dataset        & D1 Out. [\%]   & D2 Out. [\%]   & OF Out. [\%]   & SF Out. [\%]   \\
\hline
 ACOSF \cite{li2021two_acosf}            & KITTI - test  & 3.58 	       & 5.31 	       & 5.79 	 	   & 7.90          \\
 DRISF \cite{ma2019deep}            & KITTI - test  & 2.55 	       & 4.04 	       & 4.73 	 	   & 6.31          \\
 RigidMask \cite{yang2021learning_rigidmask}        & KITTI - test  & 1.89          & 3.23          & 3.50          & 4.89          \\
 RAFT-3D  \cite{teed2021raft3d}        & KITTI - test  & 1.81          & 3.67          & 4.29          & 5.77          \\
 CamLiFlow  \cite{liu2021camliflow}          & KITTI - test  & 1.81          & 2.95          & 3.10          & \textbf{4.43}          \\
 \methodname{} (ours.)     & KITTI - test  & 1.65          & 3.11          & 4.11          & 5.32          \\
 \hline
  Warped Scene Flow & FT3D - test    & 2.35          & 16.19         & 9.43          & 19.09         \\
 RigidMask         & FT3D - test    & 2.35          & 6.98          & 15.42         & 15.49         \\
 RAFT-3D           & FT3D - test    & 2.35          & 4.40          & 8.47          & \textbf{8.26}          \\
 \methodname{} (ours.)     & FT3D - test    & 2.35          & 4.86          & 8.76          & 8.73          \\
 \hline
\end{tabular}
\end{table}

\paragraph{Segmentation} For the segmentation evaluation, we retrieve a one-to-one matching between predicted and ground truth objects with the Hungarian method and report the accuracy, \ie the ratio of correctly assigned pixels. In addition to the accuracy, we report the average number of extracted objects per frame.

\begin{figure}
    \centering
    \includegraphics[scale=0.15]{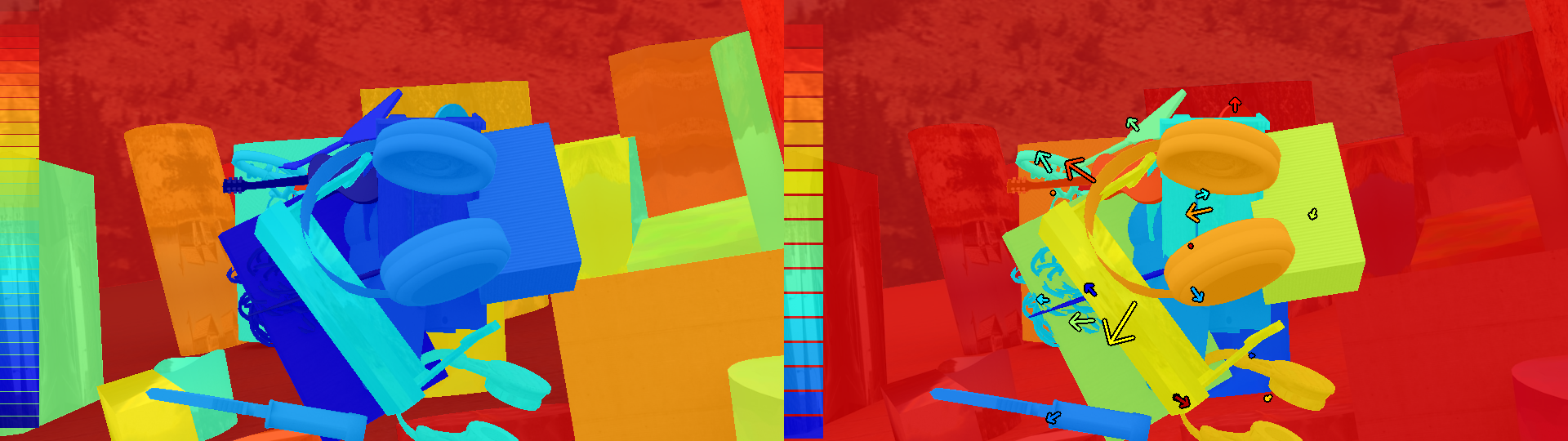}
    \caption{Based on the segmentation of objects from the FlyingThings3D dataset, illustrated on the left side, the segmentation for evaluation, shown on the right side, is retrieved. To achieve this, we fuse objects with similar $SE(3)$-motion.}
    \label{fig:eval_seg_gt_ft3d}
\end{figure}

In Table \ref{tab:eval_seg} the results are listed for the FlyingThings3D dataset. The original ground truth segmentation can not be directly used, as it splits the background into multiple objects even though they have the same $SE(3)$-motion (Fig.~\ref{fig:eval_seg_gt_ft3d} left). To resolve this, we fuse objects that have a relative pose error, as defined in Equation \ref{eq:rpe_transl} and \ref{eq:rpe_rot}, below a certain threshold (Figure \ref{fig:eval_seg_gt_ft3d} right).

\begin{table}[]
    \centering
    \caption{Results for segmenting frames into moving objects. Metrics are the segmentation accuracy and the average objects count in each frame.}
    \label{tab:eval_seg}
\begin{tabular}{llrr}
\hline
 Method        & Dataset        & Segmentation Acc. [\%]   & Objects Count [\#]   \\
\hline
 RigidMask     & FT3D - test    & 80.71                   & 16.32               \\
 \methodname{} (ours.) & FT3D - test    & \textbf{83.30}                   & 7.04                \\
\hline
\end{tabular}

\end{table}

\paragraph{Odometry}
We evaluate the odometry with the relative pose error, which has a translational part $RPE_{transl}$ and a rotational part $RPE_{rot}$. These are computed from the relative transformation $T_{rel}$ between the ground truth transformation ${}^{t_1}\hat{T}_{t_0}$ and the estimated transformation ${}^{t_1}{T}_{t_0}$, which is defined as follows:
\begin{gather}
    T_{rel} = {}^{t_1}\hat{T}_{t_0}^{-1} \: {}^{t_1}{T}_{t_0} \\
    T_{rel} = \left[\begin{array}{cc}
         R_{rel} &  t_{rel}\\
         0       &  1
    \end{array}\right].
\end{gather}
The translational and rotational relative pose errors $RPE_{transl}$ and $RPE_{rot}$ are computed as follows:
\begin{gather}
    \label{eq:rpe_transl}
    RPE_{transl} = \frac{\norm{t_{rel}}}{t_{1} - t_{0}} \text{  in  } \frac{m}{s} \\
    \label{eq:rpe_rot}
    RPE_{rot} = \frac{\norm{w{(R_{rel})}}}{t_{1} - t_{0}} \frac{360}{2 \pi} \text{  in  } \frac{deg}{s},
\end{gather}
with $w{(R_{rel})}$ being the axis-angle representation of the rotation.
We report the results on FlyingThings3D and TUM RGB-D in Table \ref{tab:eval_odo_rpe}.
\begin{table}[]
    \centering
    \caption{Results for odometry estimation on FlyingThings3D and TUM RGB-D using the translation and rotational relative pose errors $RPE_{transl}$ and $RPE_{rot}$.}
    \label{tab:eval_odo_rpe}
\begin{tabular}{llrr}
\hline
 Method        & Dataset        & RPE transl. [m/s]   & RPE rot. [deg/s]   \\
\hline

 Static        & FT3D - test    & 0.364               & 2.472              \\
 RigidMask     & FT3D - test    & 0.082               & 0.174              \\
 \methodname{} (ours.) & FT3D - test    & \textbf{0.025}               & \textbf{0.099}              \\
 \hline
 Static        & TUM FR3        & 0.156               & 18.167             \\
 RigidMask     & TUM FR3        & 0.281               & 4.345              \\
 \methodname{} (ours.) & TUM FR3        & \textbf{0.090}               & \textbf{3.535}              \\
 \hline
\end{tabular}

\end{table}

\paragraph{Runtime}
We report average runtimes of \methodname{} and the baselines in Table \ref{tab:eval_sota_runtime}. The runtimes were measured on a single Nvidia GeForce RTX 2080Ti.
\begin{table}[]
    \centering
    \caption{Runtimes for different approaches on FlyingThings3D, KITTI-2015, and TUM RGB-D. If depth and optical flow are estimated separately and the runtime is known, it is listed. Runtimes in red are from the original authors on different hardware.}
    \label{tab:eval_sota_runtime}
\begin{tabular}{llrrr}
\hline
 Method        & Dataset        & Depth [s]  & Optical Flow [s]     & Total [s]   \\
  \hline
  
 ACOSF     & KITTI - test  & -   & -                & \textcolor{red}{300.00}           \\
 DRISF     & KITTI - test  & -   & -                & \textcolor{red}{0.75}          \\
 RigidMask     & KITTI - test  & 1.46   & -        & 4.90          \\
 RAFT-3D       & KITTI - test  & 1.44   & -        & 2.73          \\
 CAMLiFlow       & KITTI - test  & -   & -        & \textcolor{red}{1.20}          \\
 \methodname{} (ours.) & KITTI - test  & 1.43   & 0.42     & 2.84          \\
\hline
 RigidMask     & FT3D - test    & 1.60   & -                  & 8.54          \\
 RAFT-3D       & FT3D - test    & 1.58   & -                  & 2.92          \\
 \methodname{} (ours.) & FT3D - test    & 1.58   & 0.40               & 3.79          \\
 \hline
 RigidMask     & TUM FR3        & 0.23   & -               & 2.34          \\
 RAFT-3D       & TUM FR3        & 0.23   & -               & 1.15          \\
 \methodname{} (ours.) & TUM FR3        & 0.23   & 0.36            & 2.29          \\
\hline
\end{tabular}
\end{table}

\section{Discussion}
\begin{figure}[!ht]
    \centering
    \includegraphics[scale=0.095]{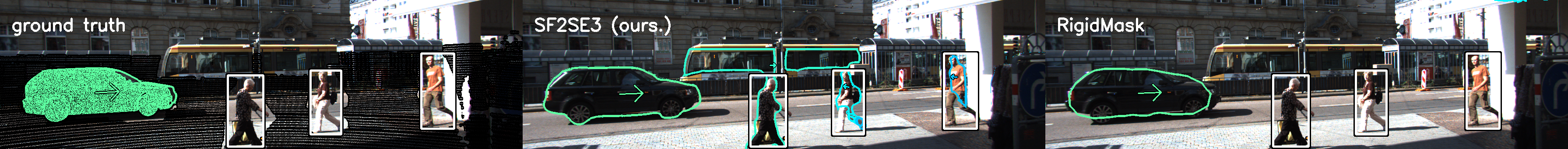}
    \caption{\textbf{Segmenting pedestrians on KITTI-2015}: ground truth, estimate from \methodname{}, estimate from RigidMask (left to right). In contrast to RigidMask, the proposed \methodname{} approach detects the pedestrians (marked with white bounding boxes) because it is not visually fine-tuned for cars. This problem of RigidMask is not reflected in the quantitative results because ground truth is missing for these points.}
    \label{fig:eval_seg_fail_pedestrians}
\end{figure}

Our method performs on par with state-of-the-art methods of the KITTI-2015 scene flow benchmark, achieving a scene flow outlier rate similar to RigidMask (\isworse{$+0.43 \%$}), CamLiFlow (\isworse{$+0.89 \%$}) and RAFT-3D (\isbetter{$-0.45 \%$}). Further, on FlyingThings3D it achieves similar scene flow performance as the pointwise method RAFT-3D (\isworse{$-0.47 \%$}) and outperforms RigidMask significantly (\isbetter{$-6.76 \%$}) while also achieving an higher segmentation accuracy (\isbetter{$+2.59 \%$}). 
In contrast to RigidMask and others, our method generalizes better because supervision is only applied for estimating optical flow and depth. Therefore, we detect the pedestrians in Figure~\ref{fig:eval_seg_fail_pedestrians}. 

\begin{figure}[!h]
    \centering
    \includegraphics[scale=0.12]{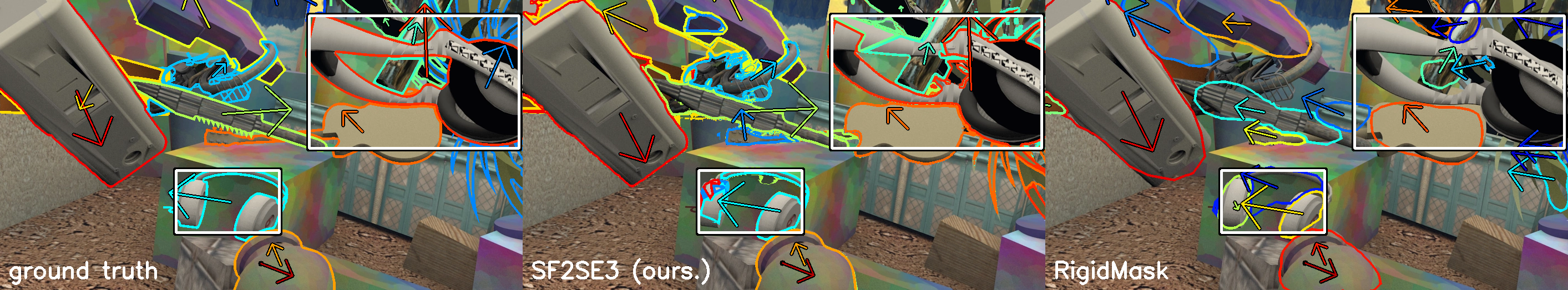}
    \caption{\textbf{Segmenting objects with complex shapes on FlyingThings3D}: ground truth, estimate from \methodname{}, estimate from RigidMask (left to right). RigidMask over-segments the semi-circular shaped headphones on the bottom and misses the headphones on the right (marked with white bounding boxes).}
    \label{fig:eval_seg_fail_polarmask_overseg_noobject}
\end{figure}

Moreover, our representation does not geometrically restrict object shapes. Thus, we are able to fit objects with complex shapes, as shown in Figure~\ref{fig:eval_seg_fail_polarmask_overseg_noobject}.

Compared to ACOSF, which is the most accurate method in scene flow on KITTI-2015 that estimates segmentation and takes no assumptions about object shapes, our method reduces the scene flow outlier percentage by (\isbetter{$-2.58 \%$}) and the runtime from 300 seconds to 2.84 seconds. Furthermore, we expect ACOSF to perform even worse in case of more objects, e.g. in FlyingThings3D, as it uses random sampling for retrieving initial $SE(3)$-motions and assumes a fixed number of objects.

\section{Conclusion}

We have proposed \methodname{}: a novel method that builds on top of state-of-the-art optical flow and disparity networks to estimate scene flow, segmentation, and odometry. In our evaluation on KITTI-2015, FlyingThings3D and TUM RGB-D, \methodname{} shows better performance than the state of the art in segmentation and odometry, while achieving comparative results for scene flow estimation.

\subsubsection*{Acknowledgment}
The research leading to these results is funded by the German Federal
Ministry for Economic Affairs and Climate Action within the project ``KI
Delta Learning'' (Förderkennzeichen 19A19013N). The authors would like
to thank the consortium for the successful cooperation.

\bibliographystyle{splncs04}
\bibliography{egbib}

\end{document}